\documentclass[10pt,twocolumn,letterpaper]{article}

\usepackage{iccv}              

\usepackage{multirow}   
\usepackage{booktabs}   
\usepackage{array}      
\usepackage{amsmath}    
\usepackage{tabularx}

\usepackage{graphicx}    
\usepackage{caption}     
\usepackage{ragged2e}    
\usepackage{adjustbox}   
\usepackage{placeins}

\usepackage{ulem}

\definecolor{iccvblue}{rgb}{0.21,0.49,0.74}
\usepackage[pagebackref,breaklinks,colorlinks,allcolors=iccvblue]{hyperref}

\def\paperID{*****} 
\def\confName{ICCV}
\def\confYear{2025}

\title{RainFusion2.0: Temporal-Spatial Awareness and Hardware-Efficient  Block-wise Sparse Attention } 

\makeatletter
\def\thanks#1{\protected@xdef\@thanks{\@thanks \protect\footnotetext{#1}}}
\makeatother
\author{
Aiyue Chen$^{*1}$\\
{\tt\small chenaiyue@huawei.com}
\and
Yaofu Liu$^{*1,2}$\\
{\tt\small yliuls@connect.ust.hk}
\and
Junjian Huang$^{*1}$\\
{\tt\small huangjunjian3@huawei.com}
\and
Guang Lian$^{*1}$\\
{\tt\small lianguang@huawei.com}
\and
Yiwu Yao$^{*1}$\\
{\tt\small yiwuyao@pku.edu.cn}
\and
Wangli Lan$^{*1}$\\
{\tt\small lanwangli@huawei.com}
\and
Jing Lin$^{*1}$\\
{\tt\small linjing28@huawei.com}
\and
Zhixin Ma$^{*1}$\\
{\tt\small mazhixin7@huawei.com}
\and
Tingting Zhou$^{*1}$\\
{\tt\small zhoutingting13@huawei.com}
\thanks{$^*$All authors contribute equally to this work as the first authors.}
\thanks{$^\dagger$Corresponding author: Harry Yang. E-mail: yangharry@ust.hk}
\thanks{$^1$ Huawei Technologies Co., Ltd \quad $^2$The Hong Kong University of Science and Technology}
}

\begin{document}
\maketitle
\begin{abstract}
In video and image generation tasks, Diffusion Transformer (DiT) models incur extremely high computational costs due to attention mechanisms, which limits their practical applications. Furthermore, with hardware advancements, a wide range of devices besides graphics processing unit (GPU), such as application-specific integrated circuit (ASIC), have been increasingly adopted for model inference. Sparse attention, which leverages the inherent sparsity of attention by skipping computations for insignificant tokens, is an effective approach to mitigate computational costs. However, existing sparse attention methods have two critical limitations: the overhead of sparse pattern prediction and \textbf{the lack of hardware generality}, as most of these methods are designed for GPU. To address these challenges, this study proposes \underline{RainFusion2.0}, which aims to develop an online adaptive, hardware-efficient, and low-overhead sparse attention mechanism to accelerate both video and image generative models, with robust performance across diverse hardware platforms. \textbf{Key technical insights} include: (1) leveraging block-wise mean values as representative tokens for sparse mask prediction; (2) implementing spatiotemporal-aware token permutation; and (3) introducing a first-frame sink mechanism specifically designed for video generation scenarios. Experimental results demonstrate that RainFusion2.0 can achieve 80\% sparsity while achieving an end-to-end speedup of 1.5~1.8× without compromising video quality. Moreover, RainFusion2.0 demonstrates effectiveness across various generative models and validates its \textbf{generalization across diverse hardware platforms}.
\end{abstract}
    
\section{Introduction}
\label{sec:intro}

Diffusion Transformer (DiT) models have exhibited outstanding performance in visual generation tasks. However, their tremendously high computational cost limits the practical application of DiT models. Among these costs, the major component stems from the Attention mechanism, whose computation scales exponentially with token length—ranging from 10K to 80K in current state-of-the-art DiT models. As research advances, it has been revealed that the Attention mechanism possesses inherent sparsity: the softmax operation in Attention enables a minor subset of tokens to exert a dominant impact on the final output. Consequently, sparse Attention, which accelerates Attention computation by only focusing on critical tokens, has emerged as a prevalent optimization approach. Currently, sparse Attention methods can be primarily categorized into three types: (1) fixed patterns, (2) online adaptive masks, and (3) token permutations. Nevertheless, most of these existing methods suffer from certain limitations, which restrict their widespread practical adoption.

\textbf{Limitations}: \uline{1.The Dilemma of balancing Accuracy and Efficiency}. Two main factors contribute to this dilemma: (1) sparsity degree; (2) the cost of predicting sparse patterns. For fixed patterns, the overhead of prediction is relatively small. However, to achieve high accuracy, fixed-pattern methods typically require a low sparsity degree, which consequently compromises efficiency. On the other hand, online adaptive mask-based and token permutation-based methods involve substantial overhead, primarily due to the costs associated with predicting the sparse mask and the token permutation layout. \uline{2.Lack of Device Universality}. A major shortcoming of existing works is their neglect of universality across various devices. Most prediction methods are specifically designed for Graphics Processing Units (GPUs) and are not universally applicable to other Artificial Intelligence (AI) devices such as Application Specific Integrated Circuit(ASIC). Specifically, the prediction process incurs unacceptably long time costs on AI devices other than GPUs.

\textbf{Goal}: We aim to design an online adaptive, hardware-efficient, and overhead-free sparse attention mechanism to accelerate video and image generation models.

\textbf{Key Idea}: \uline{1.Adaptive and Efficient Design}. We partition tokens into different blocks and use the mean value of each block as the representative token for predicting the sparse mask. \uline{2.Spatiotemporal-aware permutation}. To enhance the similarity within each block, we adopt a spatiotemporal-aware permutation strategy. Based on the inherent spatial relationships of videos or images, tokens are divided into three-dimensional (3D) or two-dimensional (2D) windows. Tokens within each window are arranged adjacently and then flattened window by window. \uline{3.First Frame Sink for Video Generation}. For video generation models, we fix the attention computation relationship with the tokens of the first frame. This fixed pattern is termed "First Frame Sink". \uline{4.Hardware Generality}. Most importantly, our proposed method not only features low overhead but also achieves generality across vary types of AI devices.

\section{Related Work}
\label{sec:formatting}


Sparse attention mechanisms have garnered significant research interest within the domain of video and image generative diffusion models. This work categorizes these advancements into three primary paradigms:

(1) \textbf{Fixed Sparse Patterns}: Sparse attention with fixed patterns typically employs a predetermined set of sparse masks. The model selects one of these masks for use. For instance, Sliding Tile Attention (STA)\cite{STA}  precomputes a set of sliding window masks. Leveraging a small number of prompts, STA matches an optimal pattern for each layer and each head within the Denoising Diffusion Transformer (DiT). During the generation of new videos, these pre-matched masks are directly used. Similarly, Sparse Video Generation (SVG)\cite{SVG}  designs two distinct pattern types: temporal and spatial masks. When computing attention, SVG dynamically checks which mask is most suitable for the current attention head. Rainfusion\cite{rainfusion} extends this concept by introducing three mask types: temporal, spatial, and textural. Analogous to SVG, Rainfusion dynamically assigns the most appropriate mask to each attention head.

(2) \textbf{Online Adaptive Patterns}: This category of sparse attention does not rely on fixed masks. Instead, tokens are generally partitioned into blocks, and the model determines online which block tokens will participate in the attention computation. SparseAttention \cite{zhang2025spargeattention} calculates the mean for each block and uses these block means to derive attention scores. It then selects the top-k blocks such that their cumulative distribution function (CDF) meets a specific threshold. Concurrently, it computes the cosine similarity between tokens within each block. Adaptive Sparse Attention (AdaSpa)\cite{AdaS} observes that the sparse patterns for each layer and head in DiT models remain stable throughout the diffusion steps. Consequently, in the initial timesteps, it maintains full attention and online computes the optimal sparse pattern for each layer, which is then reused in subsequent steps.

(3) \textbf{Token Permutations}: This approach focuses on rearranging similar tokens to be adjacent, thereby enabling effective sparsification. PAROAttention\cite{zhao2025paroattention} primarily explores permuting video tokens across different dimensions, such as frame, height, and width, resulting in six possible orderings. Sparse Video Generation 2 (SVG2)\cite{SVG2} employs K-means clustering to group similar tokens together. It then uses the centroid of each cluster as its representative. Based on these centroids, SVG2 computes the attention scores between clusters to determine which inter-cluster attention computations can be strategically omitted.

\label{sec:method}

\begin{figure*}[t]
    \centering
    \includegraphics[width=1\linewidth]{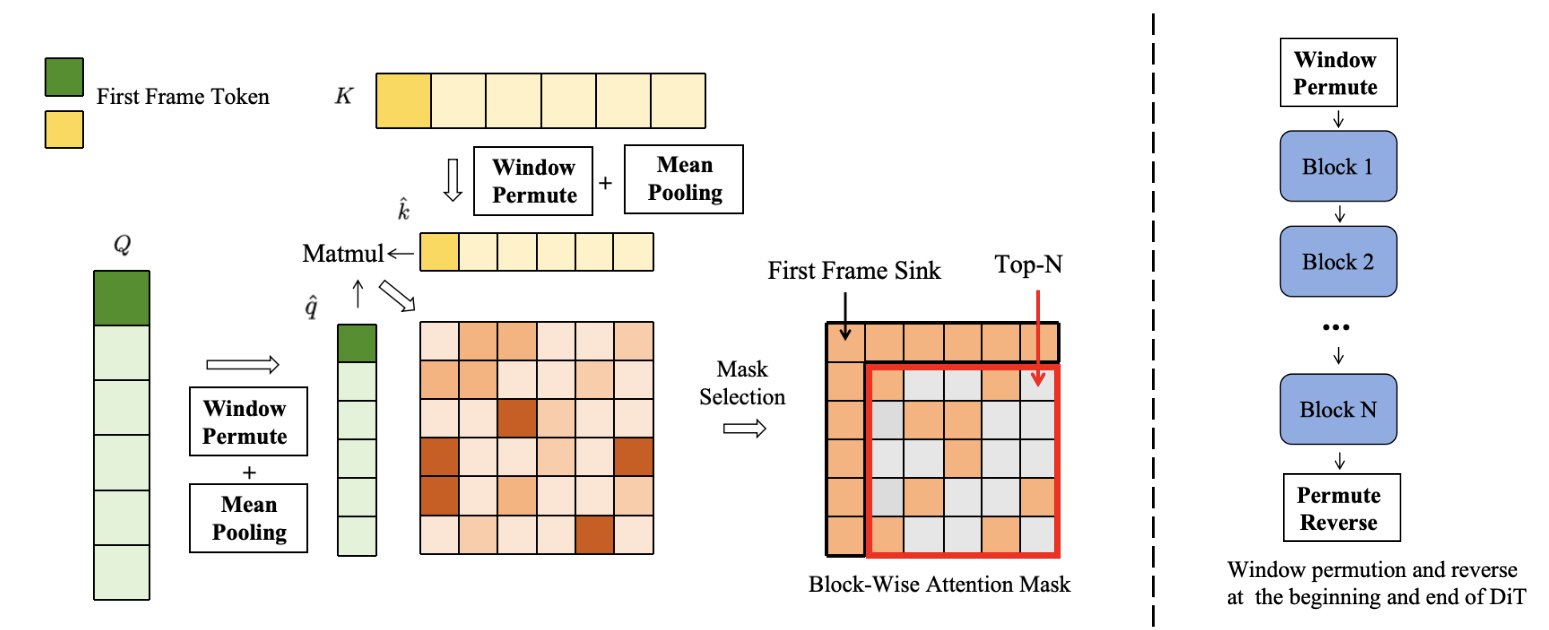}
    \caption{Workflow of RainFusion2.0 }
    \label{fig:workflow}
\end{figure*}

\section{Method}


\subsection{Blockwise Sparse}

Flash attention is the backbone of the modern transformer model. RainFusion2 adopts flash attention into the sparse mode. Consider the attention operation: given the query, key, value matrices $Q, K, V \in \mathbb{R}^{N \times d}$, we have $S = QK^\top / \sqrt{d}, \quad P = \text{Softmax}(S), \quad O = PV$ ,

where the Softmax function is defined element-wise as $\text{Softmax}(S)_{ij} = \frac{\exp(S_{ij})}{\sum_{k} \exp(S_{ik})}$.

Flash attention: Divide $Q$ into $T_q$ blocks $\{Q_i\}$ with block size $b_q$, where $Q_i \in \mathbb{R}^{b_q \times d}$ and $T_q = \left[ N / b_q \right]$. Divide $K$ and $V$ into $T_k$ blocks $\{K_i\}$ and $\{V_i\}$ respectively, with the same block size $b_k$, where $K_i \in \mathbb{R}^{b_k \times d}$, $V_i \in \mathbb{R}^{b_k \times d}$ and $T_k = \left[ N / b_k \right]$.

Subsequently, it use online softmax to compute each block of the output $O$ iteratively.  The calculation of block $O_i$  proceeds as:
\begin{align}
S_{i,j} &= Q_i K_j^\top / \sqrt{d}, \\
(m_{i,j}, \tilde{P}_{i,j}) &= \tilde{f}(m_{i,j-1}, S_{i,j}), \\
l_{i,j} &= \exp\left(m_{i,j-1} - m_{i,j}\right) l_{i,j-1} + \text{rowsum}(\tilde{P}_{i,j}), \\
O_{i,j} &= \text{diag}\left(\exp\left(m_{i,j-1} - m_{i,j}\right)\right) O_{i,j-1} + \tilde{P}_{i,j} V_j
\end{align}
where $m_{i,j}, l_{i,j} \in \mathbb{R}^{b_q \times 1}$, with initial values set to $-\infty$ and $0$ respectively. The operator $\tilde{f}(\cdot)$ is defined as follows: it computes $m_{i,j} = \max\left\{m_{i,j-1}, \text{rowmax}(S_{i,j})\right\}$ and $\tilde{P}_{i,j} = \exp\left(S_{i,j} - m_{i,j}\right)$. Finally, the block $O_i$ (the final output of this incremental process) is obtained via:
\[
O_i = \text{diag}\left(l_{i,j}\right)^{-1} O_{i,j}
\]

To accelerate attention computation and improve hardware utilization, we either skip or compute the full block-wise matrix multiplications of $Q_i K_j^\top$ and $P_{i,j} V_j$. 
We define a block-wise attention mask $M$ with dimensions $\left\lceil N/b_q \right\rceil \times \left\lceil N/b_k \right\rceil$, where each element takes a value of either 0 or 1. The sparse Flash Attention mechanism is then implemented as follows: 
\[
\text{$Q_i K_j^\top$ and $P_{i,j} V_j$ are skipped if } M_{ij} = 0.
\]

\subsection{Representative Token for Blocks}
 
To derive the block-wise attention mask \( M \), if we calculate the full attention scores \( P \) as \( Q \times K^\top \), this approach actually offers no sparsity-induced speedup.
Thus, we select representative tokens \( \hat{q}_i \) and \( \hat{k}_i \) for each block \( Q_i \) (query block) and \( K_i \) (key block), respectively. 

The core idea underlying the design of representative tokens originates from the observation that adjacent tokens in the \( Q, K \) matrix exhibit high similarity—a characteristic that is consistent across different models and also across distinct attention layers within a single model.

The representative tokens $\hat{q}_i$ and $\hat{k}_j$ are defined as follows:
\begin{align}
\hat{q} &= \{ \hat{q}_i \} = \text{mean}\left(Q_i, \text{axis}=0\right), \\
\hat{k} &= \{ \hat{k}_j \} = \text{mean}\left(K_j, \text{axis}=0\right), \\
\hat{S} &= \{ \hat{S}_{ij} \} = \{ \hat{q}_i \hat{k}_j^\top \}
\end{align}

where the matrix $\hat{S}$ has dimensions $\left\lceil N/b_q \right\rceil \times \left\lceil N/b_k \right\rceil$, and  $\{ \hat{S}_{ij} \}$ serves as an indicator to quantify the contribution of the block pair $(Q_i, K_j)$ to the overall attention scores.
Subsequently, we select the top-$n$ important blocks $\{K_j\}$ for each block $\{Q_i\}$ based on the values of $\hat{S}_{ij}$.

\begin{equation}
M_{ij} = 
\begin{cases}
1, & \text{if } j \in \left\{ j \mid \hat{S}_{ij} \in \text{TopN}\left(\hat{S}, \text{dim}=0\right) \right\}, \\
0, & \text{otherwise},
\end{cases}
\end{equation}

However, a key challenge persists in this design: Despite the high similarity between adjacent tokens, those within a single block still show considerable variability. To mitigate this issue, some studies compute the cosine similarity of tokens inside blocks \( Q_i \) and \( K_j \); yet, this approach imposes considerable computational overhead. In contrast, we employ window permutation techniques to enhance the similarity of tokens within blocks \( Q_i \) and \( K_j \).

\subsection{3D Window Permutation}

In video generation diffusion models, although tokens are flattened into 1D sequences, they inherently encode 3D physical spatiotemporal information. The length of the token sequence is given by \( N = F \cdot W \cdot H \), where \( F \), \( H \), and \( W \) denote the number of frames, spatial height, and spatial width of the video in the latent space, respectively. The tokens typically exhibit similarity to adjacent tokens in both temporal and spatial dimensions.

The default token layout in diffusion models is \( [F, H, W] \). However, this order weakens the adjacent similarity in both the temporal and spatial dimensions. Tokens that are adjacent in 3D space are shuffled into different 1D positions. Consequently, the tokens within each block are essentially non-adjacent in 3D space, thus undermining their self-similarity.

\textbf{Permutation}: To enhance the similarity of tokens within each block (i.e., \( Q_i \) and \( K_j \)), we rearrange the tokens in a 3D window-based order. The details will be released later.

\subsection{First Frame Sink}

Similar to the well-documented "attention sink" phenomenon in large language models (LLMs), where initial tokens receive high levels of attention, an analogous phenomenon exists in video generation models with 3D full attention. We refer to this analogous phenomenon in video generation as the \textit{First Frame Sink}.

Our observations across a range of video generation models reveal that the first frame exerts a substantial influence on the quality of the final generated video. Specifically, omitting the attention computation involving tokens corresponding to the first frame leads to a non-trivial degradation in the quality of the generated video. Thus, tokens in $Q$ that represent the first frame are designed to attend to all tokens in $K$, while all tokens in $Q$ are enforced to attend to tokens in $K$ that correspond to the first frame.

In the \cref{fig:workflow}, the first frame token is shown at the beginning of the sequence. However, in practice, we move the first frame token to the end of the sequence. This is because some models combine video and text tokens as multimodal input for attention computation. By grouping the first frame token with the text tokens, we can ensure they both participate in full attention.









\section{Experiment}

\begin{table*}[tp!]
 \vfill 
  \centering
  \footnotesize  
  \caption{RainFusion2 and Full Attention Comparison: Quality \& Efficiency Metrics (Wan2.2 720p)}

  \newcolumntype{Y}{>{\centering\arraybackslash}X}
  \begin{tabularx}{\linewidth}{>{\raggedright\arraybackslash}p{4cm} Y Y Y Y >{\centering\arraybackslash}p{1.5cm} >{\centering\arraybackslash}p{1cm}}
    \toprule  
    \multirow{2}{*}{Method(sparsity)} & \multicolumn{4}{c}{Quality Metrics} & \multicolumn{2}{c}{Efficiency Metrics} \\
    \cmidrule(r){2-5} \cmidrule(l){6-7}
    & Subj. Cons. $\uparrow$  & Imaging Qual. $\uparrow$ & Overall Cons. $\uparrow$ & Cosine Sim. $\uparrow$ & Latency (s) & Speedup \\
    \midrule 
    Full Attention & 0.9717 & 0.6816 & 0.2591 & & 532 & \\
    RainFusion(80\%)(w/o 3D order) & 0.9690 & 0.6791 & 0.2555 & 0.9532 & 339 & 1.57x \\
    RainFusion(90\%)(w/o 3D order) & 0.9643 & 0.6709 & 0.2555 & 0.9476 & 295 & 1.80x \\
    RainFusion(80\%)(w/ 3D order) & 0.9683 & 0.6864 & 0.2562 & 0.9514 & 339 & 1.57x \\
    \bottomrule  
  \end{tabularx}
\end{table*}

\begin{figure*}[bhtp!]
  \centering
  \setlength{\tabcolsep}{2pt} 
  \newlength{\colwidthfive}   
  \setlength{\colwidthfive}{0.15\textwidth}
  \newlength{\colwidthsix}    
  \setlength{\colwidthsix}{0.12\textwidth} 
  \newlength{\rowheight}      
  \setlength{\rowheight}{0.09\textheight}

  \begin{tabular}{>{\centering\arraybackslash}m{0.08\textwidth}  
                >{\centering\arraybackslash}m{\colwidthfive}   
                >{\centering\arraybackslash}m{\colwidthfive}|  
                >{\centering\arraybackslash}m{\colwidthfive}   
                >{\centering\arraybackslash}m{\colwidthfive}   
                >{\centering\arraybackslash}m{\colwidthfive}|  
                >{\centering\arraybackslash}m{\colwidthsix}}   
    & \multicolumn{2}{c}{\small{HunyuanVideo1.5}} & 
      \multicolumn{3}{c}{\small{Wan2.2}} & 
      \small{Qwen-image-edit} \\[4pt] 
    
    \small\textbf{Full} &
    \adjustbox{valign=c}{\includegraphics[height=\rowheight, width=\colwidthfive, keepaspectratio]{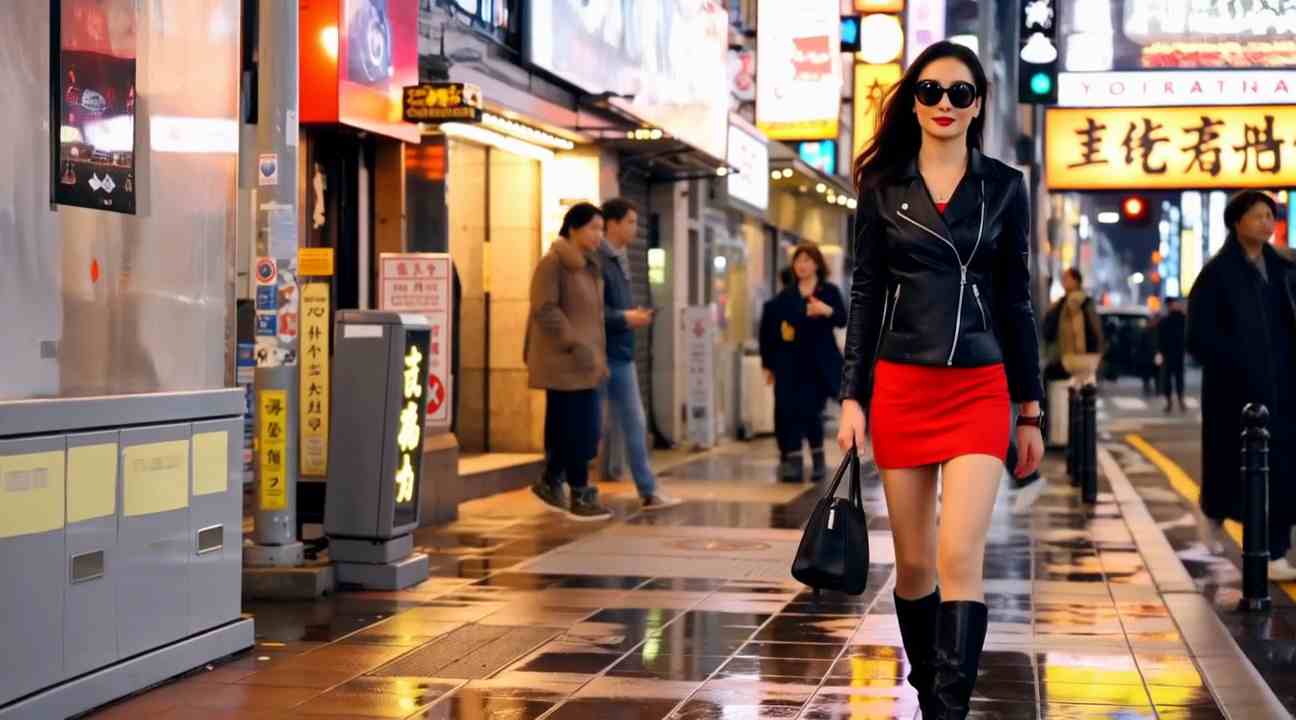}} &
    \adjustbox{valign=c}{\includegraphics[height=\rowheight, width=\colwidthfive, keepaspectratio]{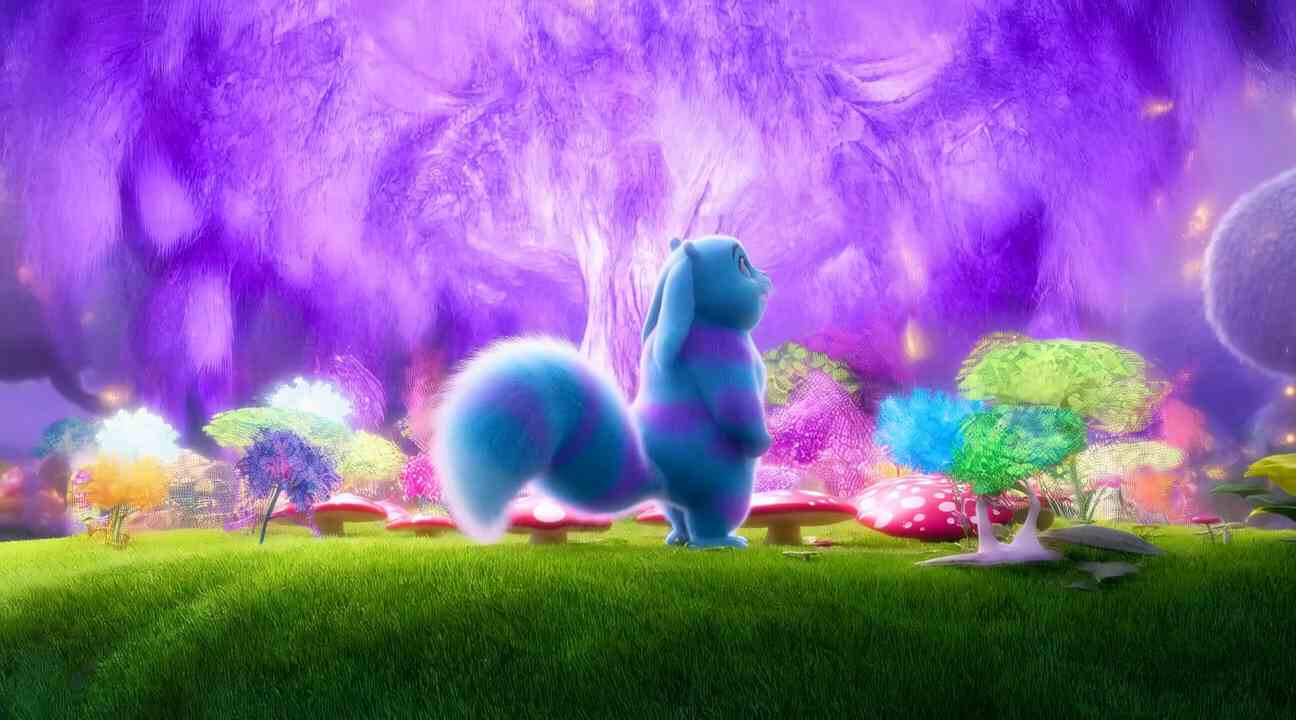}} &
    \adjustbox{valign=c}{\includegraphics[height=\rowheight, width=\colwidthfive, keepaspectratio]{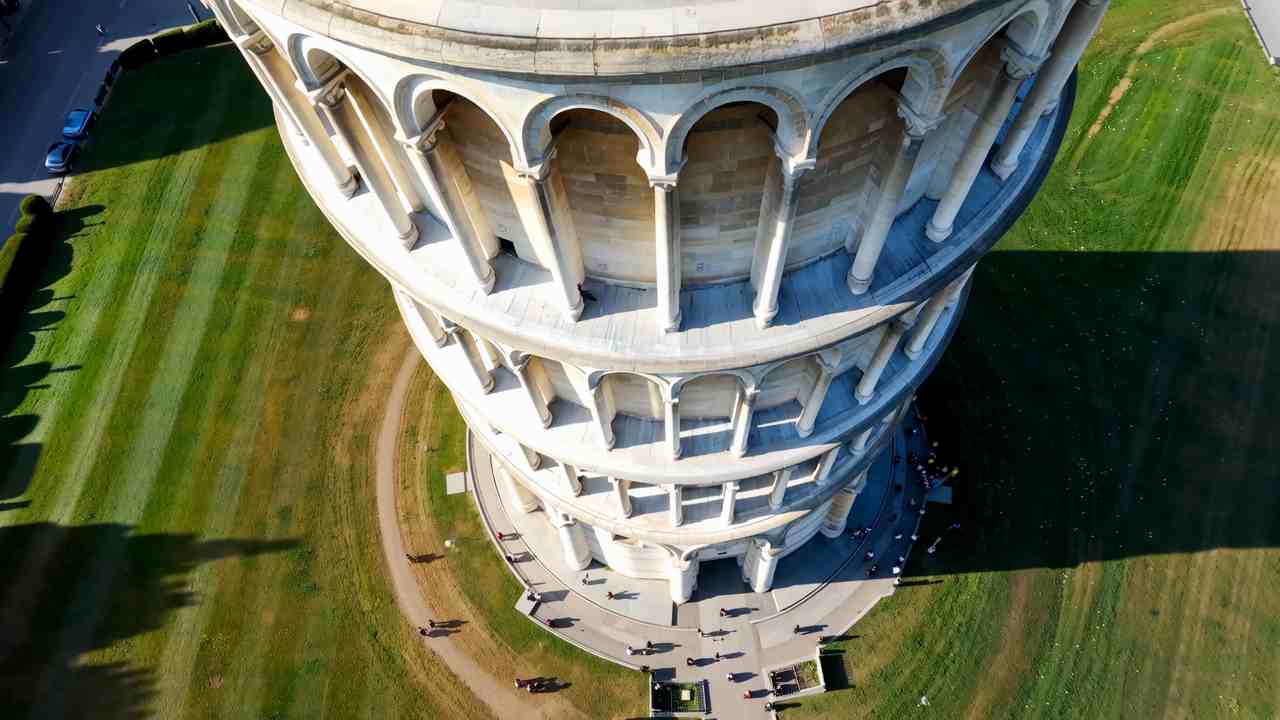}} &
    \adjustbox{valign=c}{\includegraphics[height=\rowheight, width=\colwidthfive, keepaspectratio]{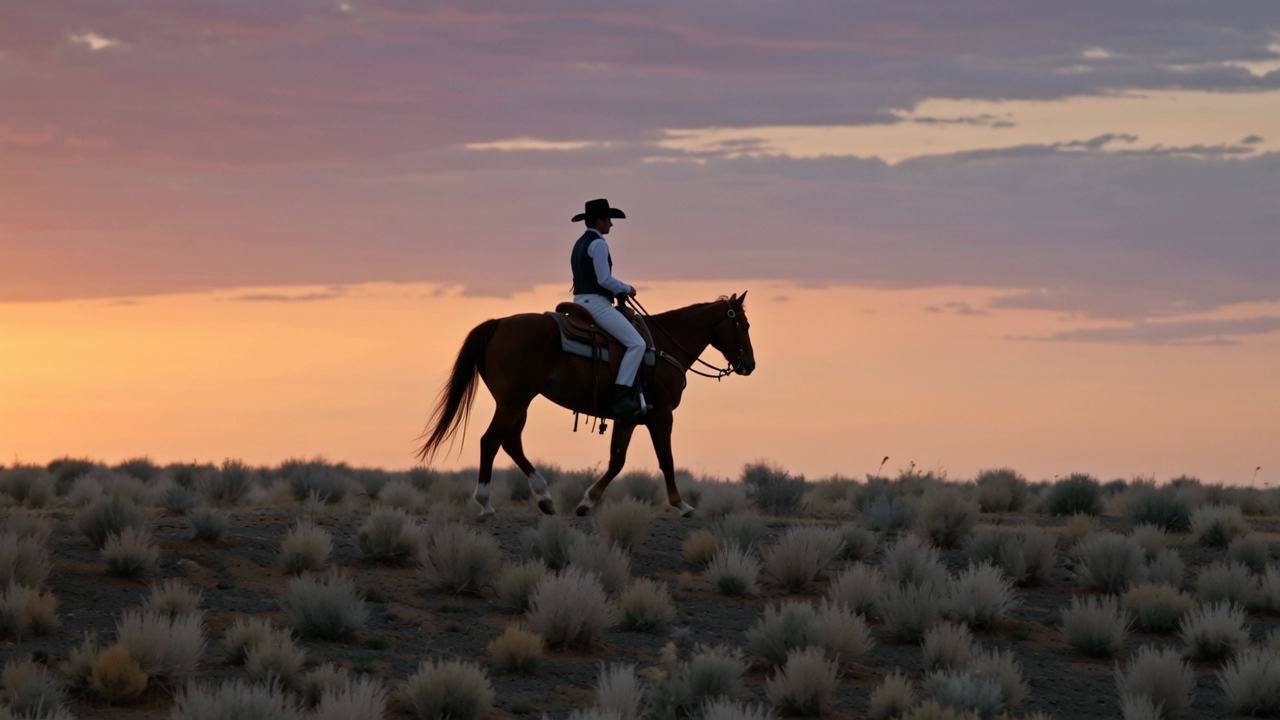}} &
    \adjustbox{valign=c}{\includegraphics[height=\rowheight, width=\colwidthfive, keepaspectratio]{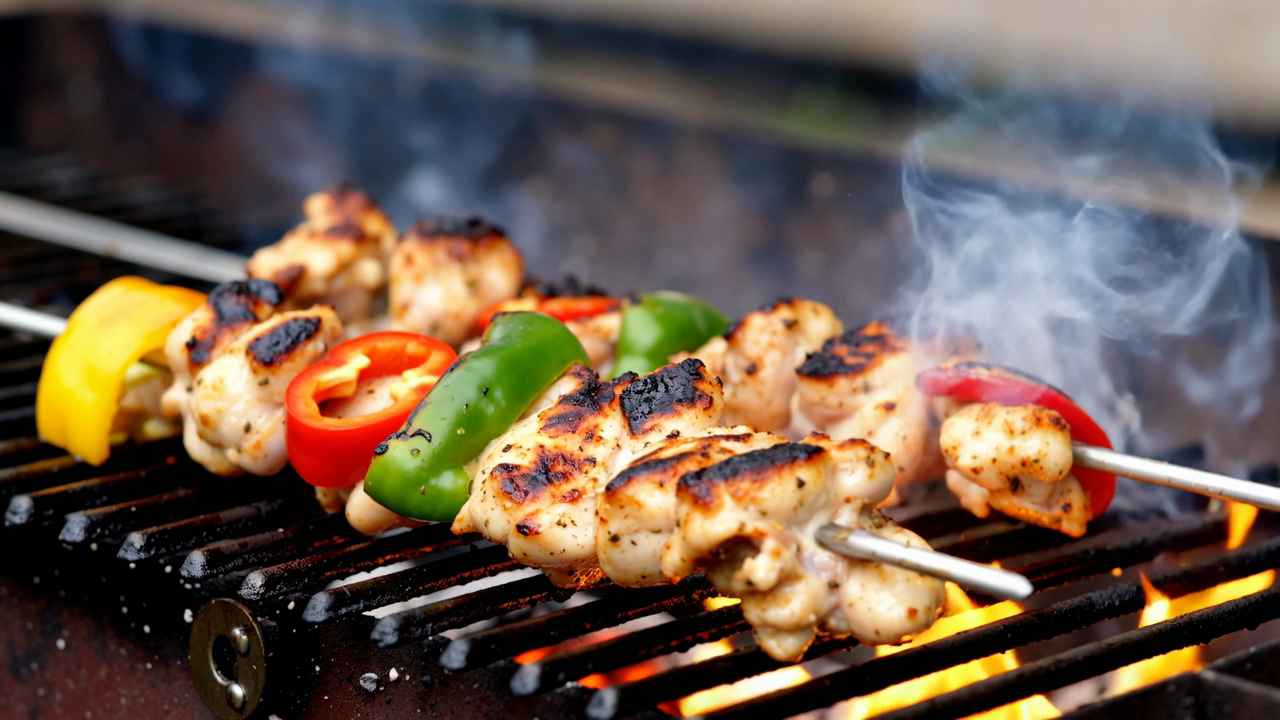}} &
    \adjustbox{valign=c, totalheight=\rowheight, max width=\colwidthsix}{\includegraphics[keepaspectratio]{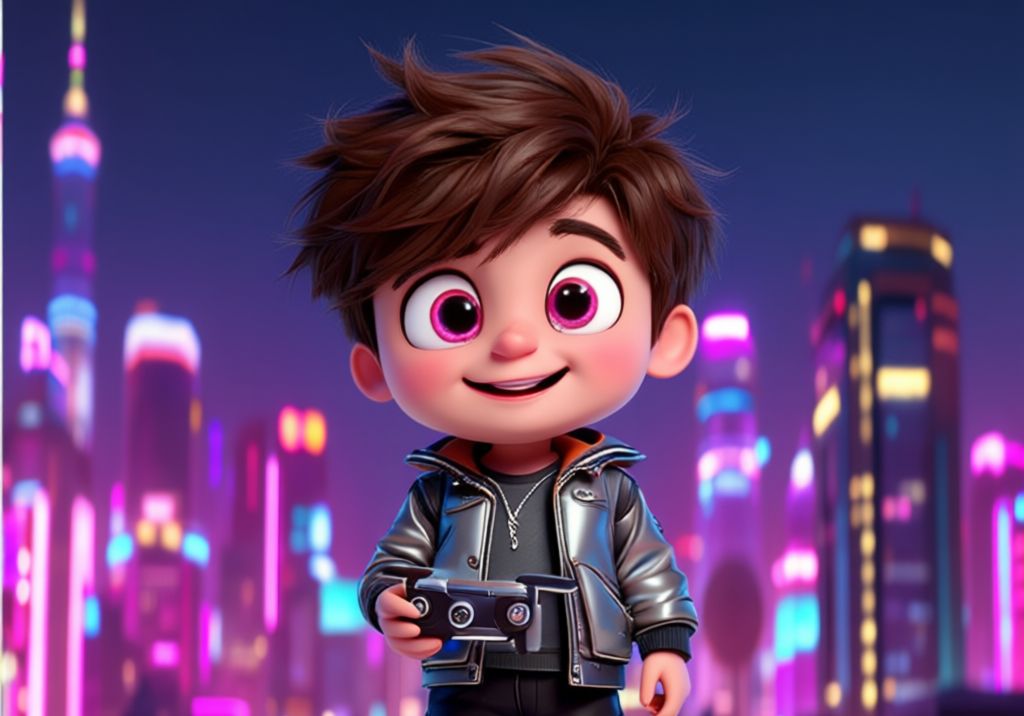}} \\[8pt]
    
    \small\textbf{RainFusion} &
    \adjustbox{valign=c}{\includegraphics[height=\rowheight, width=\colwidthfive, keepaspectratio]{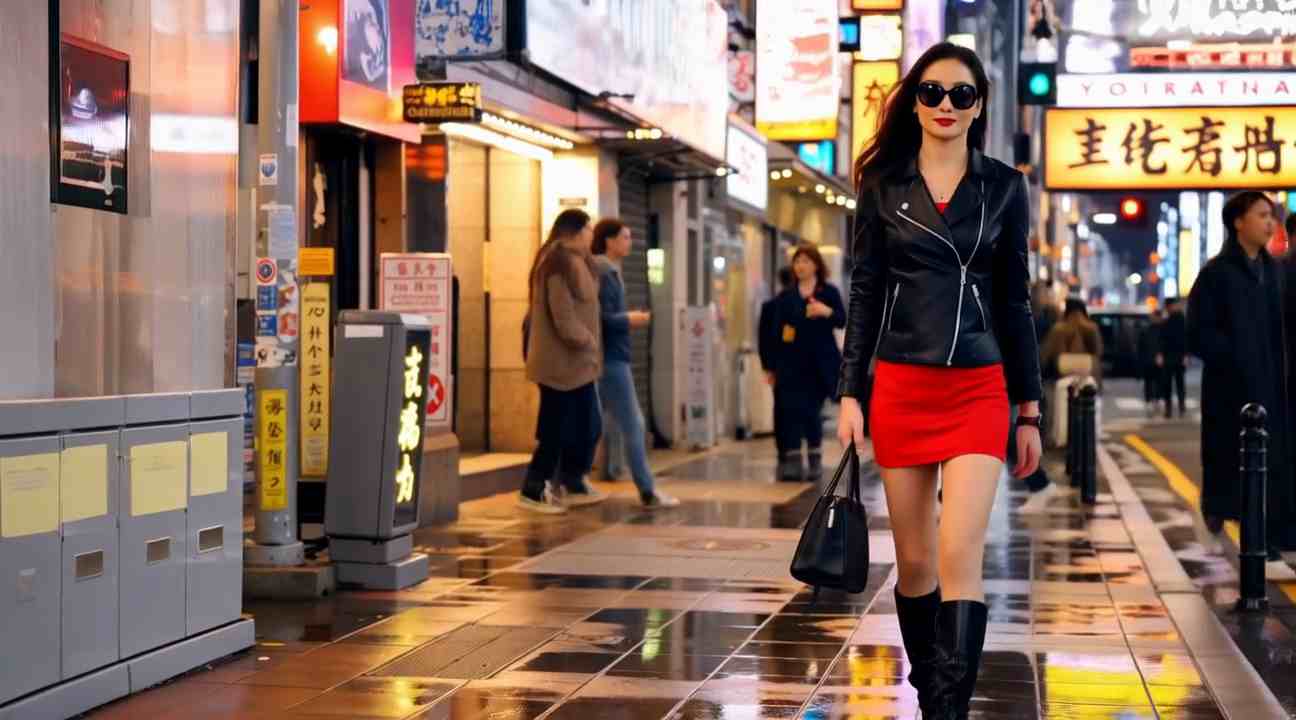}} &
    \adjustbox{valign=c}{\includegraphics[height=\rowheight, width=\colwidthfive, keepaspectratio]{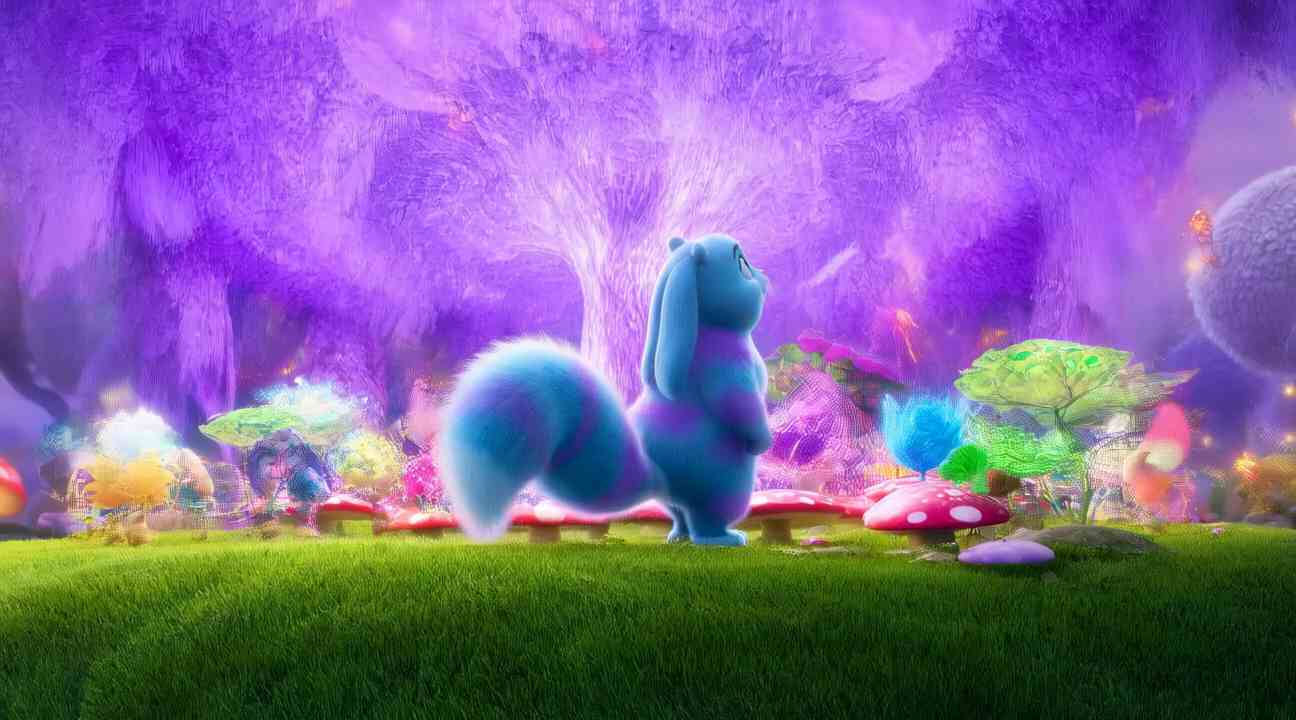}} &
    \adjustbox{valign=c}{\includegraphics[height=\rowheight, width=\colwidthfive, keepaspectratio]{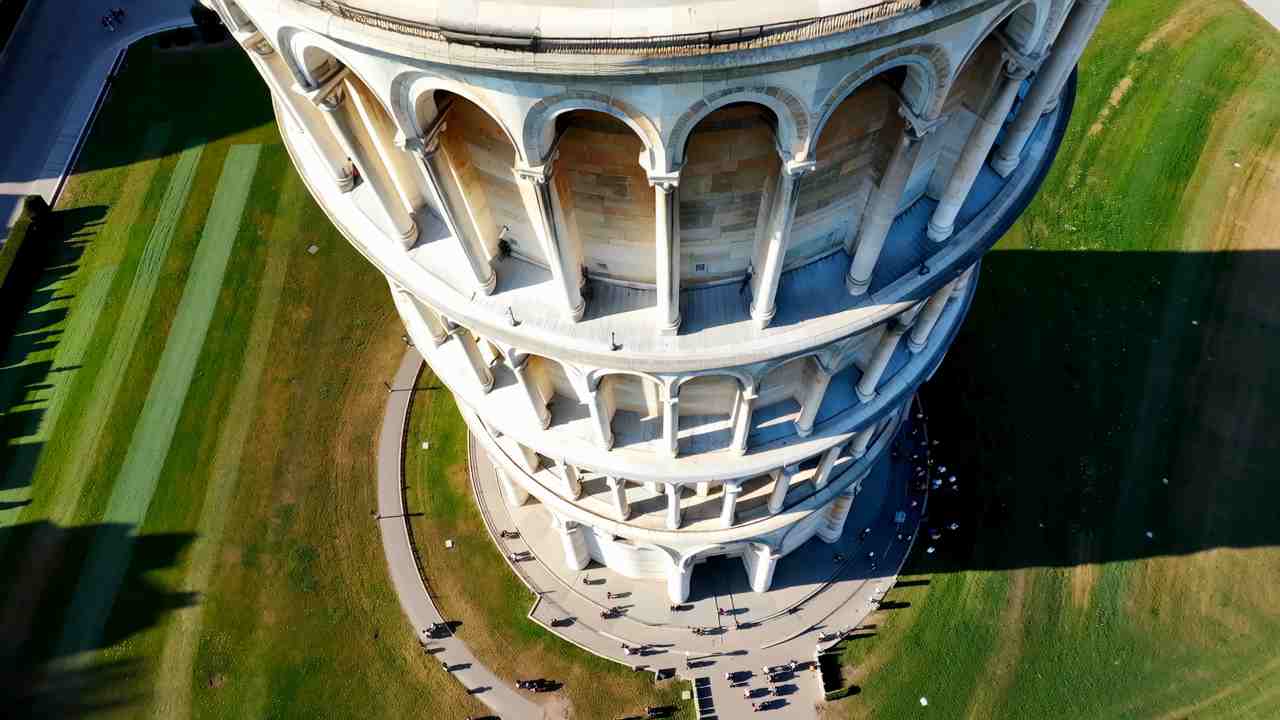}} &
    \adjustbox{valign=c}{\includegraphics[height=\rowheight, width=\colwidthfive, keepaspectratio]{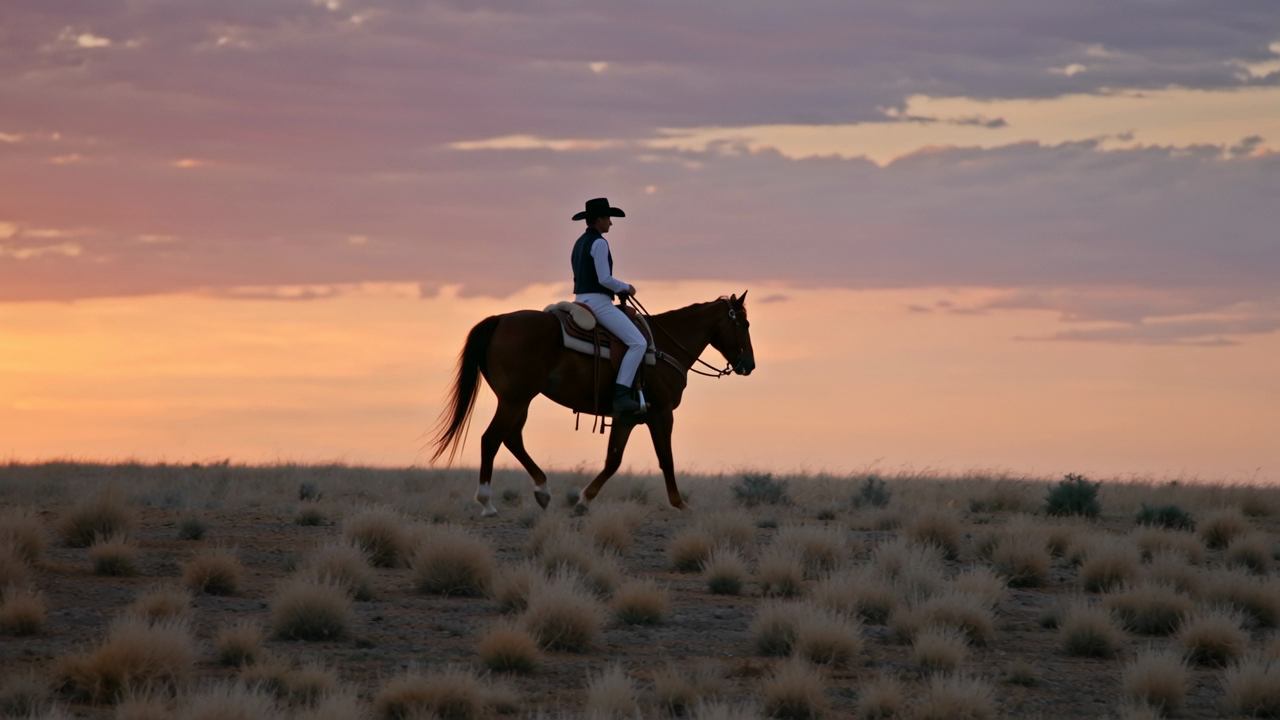}} &
    \adjustbox{valign=c}{\includegraphics[height=\rowheight, width=\colwidthfive, keepaspectratio]{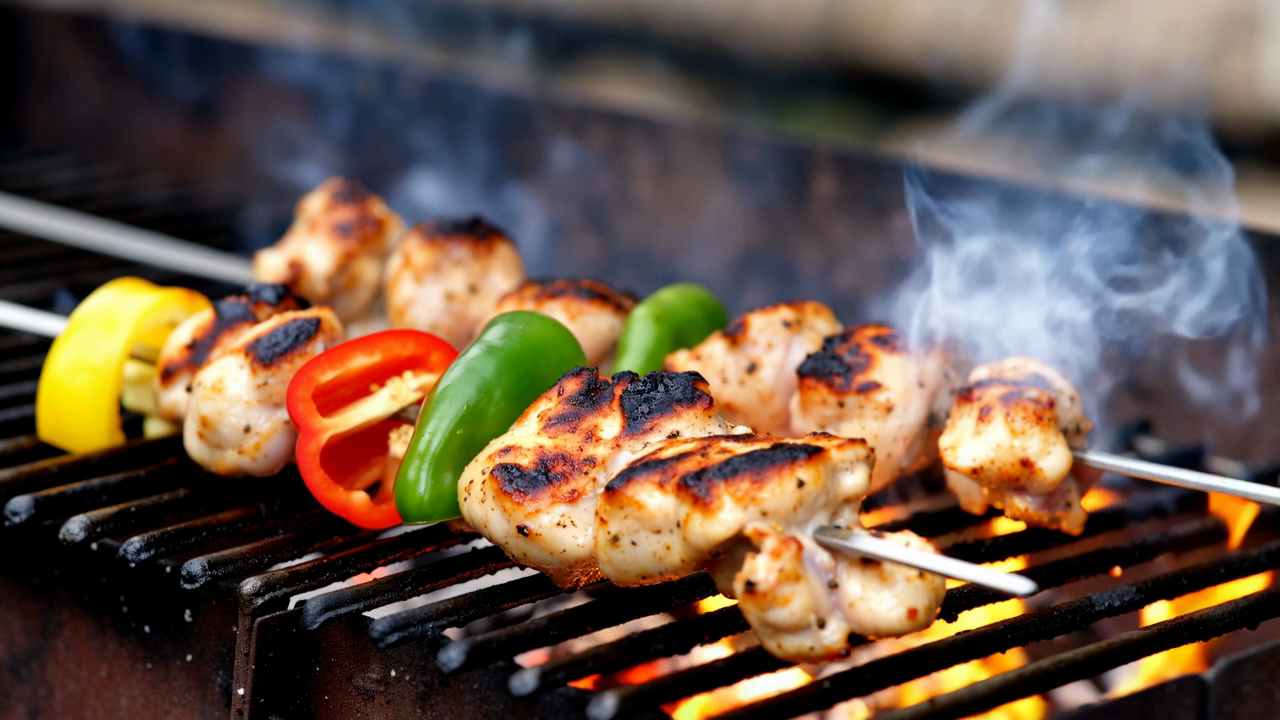}} &
    \adjustbox{valign=c, totalheight=\rowheight, max width=\colwidthsix}{\includegraphics[keepaspectratio]{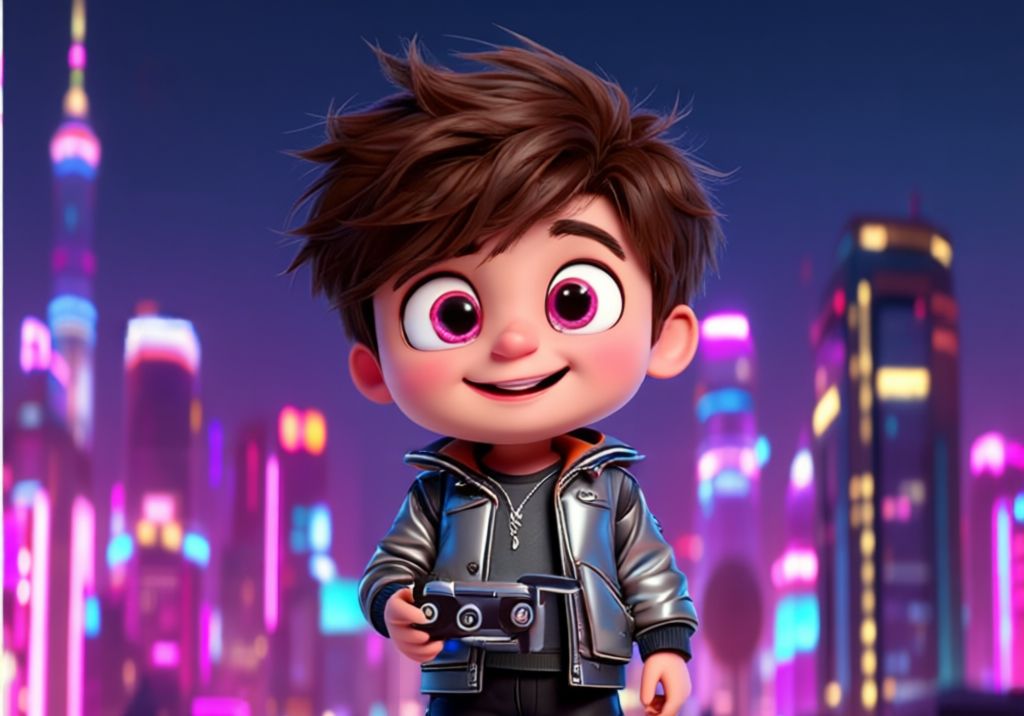}}
  \end{tabular}

  \caption{Results of RainFusion on Diffusion Models.
HunyuanVideo1.5 and Wan2.2 generate 720p videos under two configurations: full attention and RainFusion with 80\% sparsity. Qwen-image-edit generates 1024×1024 images using RainFusion with 60\% sparsity.}
  \label{fig:main3}
\end{figure*}


\begin{figure*}[htbp!]  
  \centering  
  
  \begin{tabular}{@{}c@{\hspace{0.8em}}c@{\hspace{0.8em}}c@{}}
    \textbf{(a) Full Attention} &  
    \textbf{(b) RainFusion 80\% (w/o 3D permutation)} &
    \textbf{(c) RainFusion 80\% (w/ 3D permutation)} \\
    \includegraphics[width=0.28\linewidth]{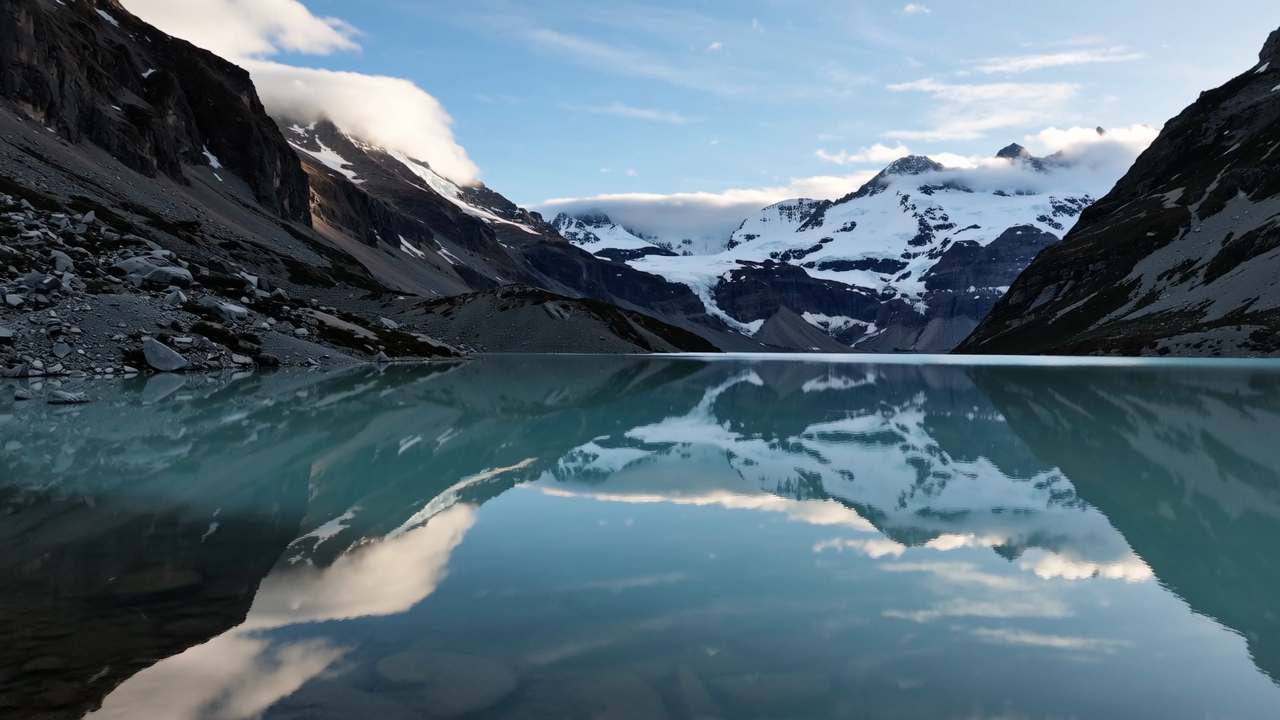} &
    \includegraphics[width=0.28\linewidth]{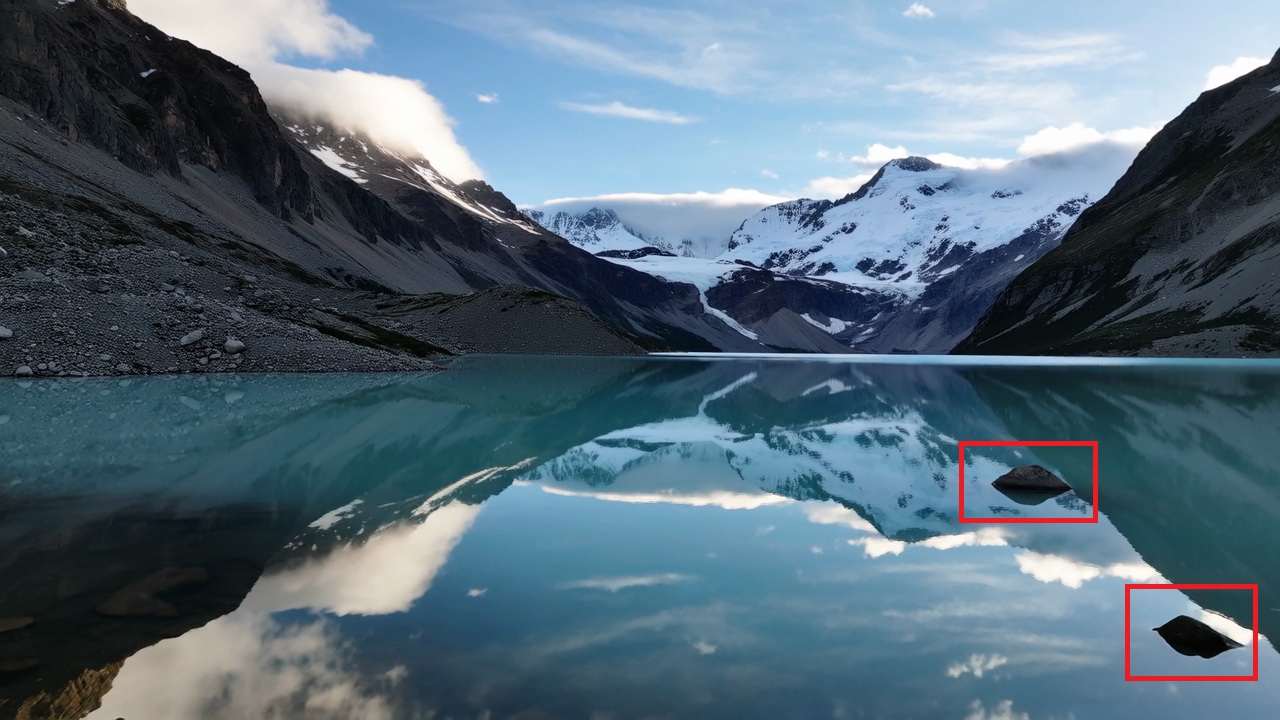} &
    \includegraphics[width=0.28\linewidth]{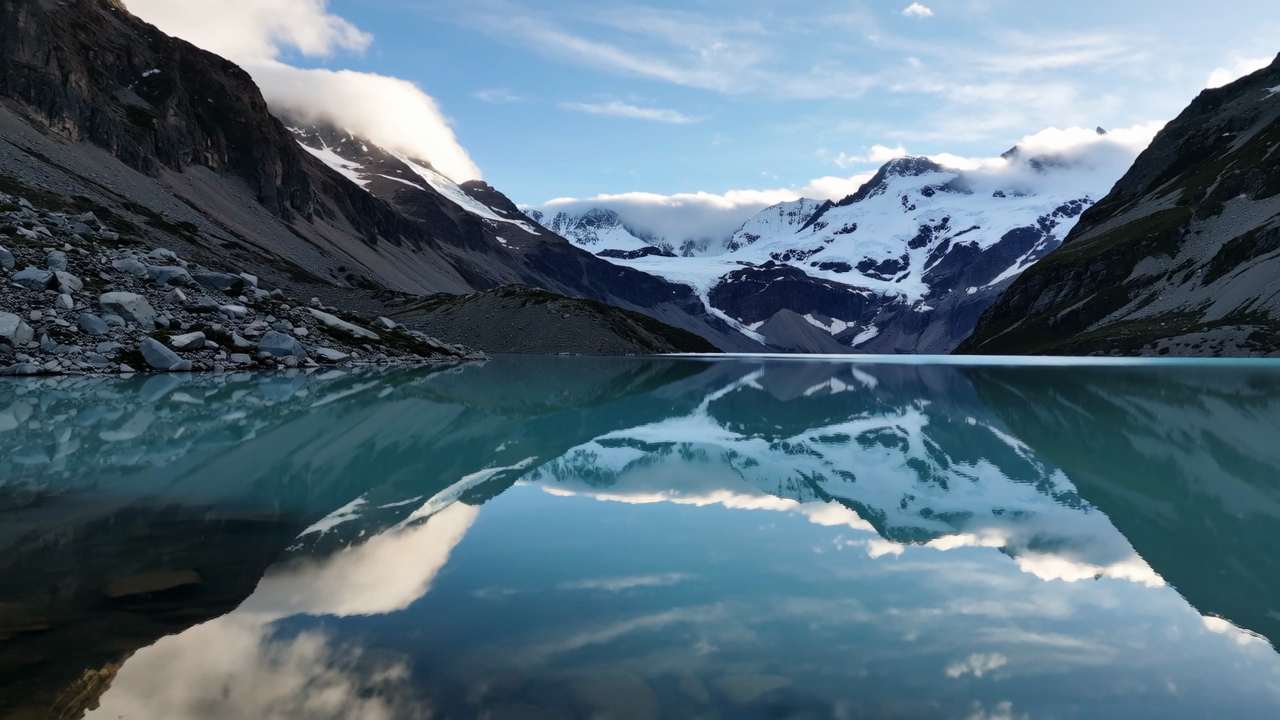} \\
  \end{tabular}
  
  \caption{Experimental results on the Wan2.2 dataset. As shown in Subfigure (b), the video generated by RainFusion (80\% sparsity, without 3D permutation) is overall comparable to that of full attention (Subfigure (a)). However, two prominent spurious rock artifacts emerge in the bottom-right corner of the video frame. In contrast, these visual artifacts are completely eliminated when 3D permutation is integrated into RainFusion (Subfigure (c)).}
  \label{fig:rainfusion_subfigs}  
\end{figure*}

\subsection{setup}
\textbf{Device}: We evaluated our method on the Neural Processing Unit (NPU), a type of Application-Specific Integrated Circuit (ASIC). As a typical and prevalent AI device distinct from Graphics Processing Units (GPUs), NPUs have been widely deployed for the inference of diffusion models. To highlight the hardware universality and efficiency of our proposed method, we conducted comprehensive performance evaluations on NPUs.

\textbf{Model}: We validated the effectiveness of RainFusion2 by applying it to various generative models, covering both video and image generation tasks.

\textit{Video Generation Models}: Experiments were conducted on Wan2.2 and Hunyuanvideo1.5. For Wan2.2, we performed tests on the 720P resolution. For Hunyuanvideo1.5, evaluations were carried out on both 480P and 720P video resolutions. Notably, Hunyuanvideo1.5 inherently incorporates a sparse attention mechanism when generating 720P videos. We thus conducted comparative experiments between RainFusion2 and this native sparse attention mechanism in terms of both accuracy and performance.

\textit{Image Generation Model}: We additionally implemented experiments on the image generation model Qwen-image-edit.

\textbf{Evaluation Metrics}: For video generation models, Vbench\cite{VBench} was primarily adopted as the core evaluation metric for video quality, which encompasses multiple critical dimensions including subjective consistency, imaging quality, and overall consistency. Furthermore, we introduced cosine similarity to characterize the overall differences between videos generated by the full attention mechanism and those generated by RainFusion2.

\textbf{Baseline}: Full Attention was employed as the fundamental baseline for comparison. We tested RainFusion2 with different sparsity levels (80\% and 90\%). Other sparse attention methods (e.g., SparseAttention and SVG2) exhibit poor compatibility with NPUs, and thus were not included in the efficiency comparison experiments.

\textbf{Perceptual Evaluation}: While Vbench metrics can reflect the overall quality of videos and cosine similarity can characterize overall differences, these metrics struggle to capture the fine-grained differences that are perceptible to the human eye. To address this limitation, we incorporated perceptual evaluation to quantitatively assess the detailed differences between videos.

\subsection{Main result}

\textit{Video Generation Models}: For Wan2.2 generating 720p videos, we achieve a speedup of 1.57× to 1.8× with a sparsity ratio of 80\% to 90\%, while maintaining nearly imperceptible quality loss (visually indistinguishable from the full attention baseline). For HunyuanVideo1.5, configuring a sparsity ratio of 80\% to 90\% also yields videos with almost no visual quality degradation. Specifically, at 80\% sparsity, HunyuanVideo1.5 achieves a 1.16× speedup for 480p video generation and a 1.28× speedup under the 720p setting.

\textit{Image Generation Model}: We also validated our method on the Qwen Image Edit model. Even with a sparsity ratio of 60\%, the model maintains consistent quality in generated images (visually aligned with the full attention baseline).

\textit{Ablation Study}: We conduct ablation experiments on the 3D window permutation. Although omitting permutation still performs well in quality metrics, and its cosine similarity is nearly identical to that of the videos generated with full attention, visual inspection of the generated videos reveals minor yet noticeable differences in temporal motion smoothness and frame details. For example, as shown in \cref{fig:rainfusion_subfigs}, videos generated without 3D window permutation are generally similar to those of full attention, but obvious flaws appear in some local areas — these differences are hard to capture with common quality metrics. In contrast, adding 3D window permutation results in videos that are visually consistent with full attention, both in overall appearance and fine-grained details.

\section{Conclusion}

In conclusion, this study has addressed the critical challenges of high computational cost in DiT and limited hardware generality in sparse attention by proposing RainFusion2.0. Through the innovative integration of block-wise representative token prediction, spatiotemporal-aware token permutation, and a first-frame sink mechanism, RainFusion2.0 achieves a remarkable balance between efficiency and performance. The experimental results demonstrate that it can achieve 80\% sparsity, leading to an end-to-end speedup of 1.5~1.8× across various video and image generative models, all while maintaining high output quality. More importantly, its hardware-efficient design allows RainFusion2.0 to work across a wide range of platforms, effectively mitigating the hardware-specific limitations of existing sparse attention methods. This work not only provides a practical and high-performance solution for accelerating DiT models but also paves the way for their broader deployment on heterogeneous computing devices. Future work will focus on exploring how to combine RainFusion2.0 with other orthogonal acceleration methods, such as quantization and distillation.
{
    \small
    \bibliographystyle{ieeenat_fullname}
    \bibliography{main}

\begin{thebibliography}{8}
\providecommand{\natexlab}[1]{#1}
\providecommand{\url}[1]{\texttt{#1}}
\expandafter\ifx\csname urlstyle\endcsname\relax
  \providecommand{\doi}[1]{doi: #1}\else
  \providecommand{\doi}{doi: \begingroup \urlstyle{rm}\Url}\fi

\bibitem[Chen et~al.(2025)]{rainfusion}
Aiyue Chen et~al.
\newblock Rainfusion: Adaptive video generation acceleration via multi-dimensional visual redundancy.
\newblock In \emph{arXiv preprint arXiv:2505.21036}, 2025.

\bibitem[Huang et~al.(2024)Huang, He, Yu, Zhang, Si, Jiang, Zhang, Wu, Jin, Chanpaisit, Wang, Chen, Wang, Lin, Qiao, and Liu]{VBench}
Ziqi Huang, Yinan He, Jiashuo Yu, Fan Zhang, Chenyang Si, Yuming Jiang, Yuanhan Zhang, Tianxing Wu, Qingyang Jin, Nattapol Chanpaisit, Yaohui Wang, Xinyuan Chen, Limin Wang, Dahua Lin, Yu Qiao, and Ziwei Liu.
\newblock Vbench: Comprehensive benchmark suite for video generative models.
\newblock In \emph{CVPR}, pages 21807--21818, 2024.

\bibitem[Xi et~al.(2025)]{SVG}
Haocheng Xi et~al.
\newblock Sparse videogen: Accelerating video diffusion transformers with spatial-temporal sparsity.
\newblock In \emph{Forty-second International Conference on Machine Learning}, 2025.

\bibitem[Xia et~al.(2025)Xia, Ling, Fu, Wang, Li, Xiao, and Cui]{AdaS}
Yifei Xia, Suhan Ling, Fangcheng Fu, Yujie Wang, Huixia Li, Xuefeng Xiao, and Bin Cui.
\newblock Training-free and adaptive sparse attention for efficient long video generation.
\newblock In \emph{Proceedings of the IEEE/CVF International Conference on Computer Vision (ICCV)}, pages 15982--15993, 2025.

\bibitem[Yang et~al.(2025)Yang, Xi, Zhao, Li, Zhang, Cai, Lin, Li, Xu, Peng, Chen, Han, Keutzer, and Stoica]{SVG2}
Shuo Yang, Haocheng Xi, Yilong Zhao, Muyang Li, Jintao Zhang, Han Cai, Yujun Lin, Xiuyu Li, Chenfeng Xu, Kelly Peng, Jianfei Chen, Song Han, Kurt Keutzer, and Ion Stoica.
\newblock Sparse videogen2: Accelerate video generation with sparse attention via semantic-aware permutation.
\newblock In \emph{The Thirty-ninth Annual Conference on Neural Information Processing Systems}, 2025.

\bibitem[Zhang et~al.(2025{\natexlab{a}})Zhang, Xiang, Huang, wei, Xi, Zhu, and Chen]{zhang2025spargeattention}
Jintao Zhang, Chendong Xiang, Haofeng Huang, Jia wei, Haocheng Xi, Jun Zhu, and Jianfei Chen.
\newblock Spargeattention: Accurate and training-free sparse attention accelerating any model inference.
\newblock In \emph{Forty-second International Conference on Machine Learning}, 2025{\natexlab{a}}.

\bibitem[Zhang et~al.(2025{\natexlab{b}})]{STA}
Peiyuan Zhang et~al.
\newblock Fast video generation with sliding tile attention.
\newblock In \emph{Forty-second International Conference on Machine Learning}, 2025{\natexlab{b}}.

\bibitem[Zhao et~al.(2025)Zhao, Hong, Yang, Xiao, Li, Ling, Xie, Chen, Zhu, Yichong, and Wang]{zhao2025paroattention}
Tianchen Zhao, Ke Hong, Xinhao Yang, Xuefeng Xiao, Huixia Li, Feng Ling, Ruiqi Xie, SiQi Chen, Hongyu Zhu, Zhang Yichong, and Yu Wang.
\newblock {PAROA}ttention: Pattern-aware reordering for efficient sparse and quantized attention in visual generation models.
\newblock In \emph{The Thirty-ninth Annual Conference on Neural Information Processing Systems}, 2025.

\end{thebibliography}
}

\end{document}